# An experimental study in Real-time Facial Emotion Recognition on new 3RL dataset

**Rahmeh Abou Zafra\*, Lana Ahmad Abdullah, Rouaa Alaraj, Rasha Albezreh, Tarek Barhoum, and Khloud Al Jallad**

*Arab International University Daraa, Syria.*

**\*Corresponding Author**
Rahmeh Abou Zafra, Arab International University. Daraa, Syria.

*Abstract*
*Although real-time facial emotion recognition is a hot topic research domain in the field of human-computer interaction, state-of-the-art available datasets still suffer from various problems, such as some unrelated photos such as document photos, unbalanced numbers of photos in each class, and misleading images that can negatively affect correct classification. The 3RL dataset was created, which contains approximately 24K images and will be publicly available, to overcome previously available dataset problems. The 3RL dataset is labelled with five basic emotions: happiness, fear, sadness, disgust, and anger. Moreover, we compared the 3RL dataset with other famous state-of-the-art datasets (FER dataset, CK+ dataset), and we applied the most commonly used algorithms in previous works, SVM and CNN. The results show a noticeable improvement in generalization on the 3RL dataset. Experiments have shown an accuracy of up to 91.4% on 3RL dataset using CNN where results on FER2013, CK+ are, respectively (approximately from 60% to 85%).*

## Introduction
Facial emotion recognition is a technology that analyses facial expressions from both static images and videos to reveal information on one's emotional state. The shape of the eyebrows, lips, nose, and chin plays an important role in determining facial expressions.

This paper proposed a new dataset to overcome these problems. Moreover, two models are applied to compare our dataset with previously available datasets.

The first model is SVM using landmark and HOG (histogram of oriented gradients) feature descriptors extracted from images. The second model is CNN using images and landmarks as features.
The paper is organized as follows: section 1 is an introduction, and section 2 is related works. Section 3 is about previously available datasets. Experiments are shown in section 4 and then setup, and finally section 6 contains conclusion and future work.

## Related Works
SVM is one of the most commonly used machine learning models in FER (Facial Expression Recognition), as in (Alshamsi and Këpuska)[1]. Alshamsi et al. used SVM for classification, COG and landmarks for feature extraction. In (Patwardhan)[2], (Youssef, Aly and Ibrahim)[3], and (Zhang, Cui and Liu)[4], geometric, kinematic and extracted features from daily behavioral patterns were used in feature extraction, whereas the SVM algorithm was used to recognize emotions and constructed a facial emotion recognition dataset that contains 84 samples. SVM and k-NN (k Nearest Neighbors) are used to classify emotions.

SVMs have proven their qualification of multiclass classification. As in (Alshamsi and Këpuska)[1], (Joseph and P. Geetha)[5], (Lucey, Cohn and Kanade)[6], (Duan and Keerthi)[7], (Chandran and Dr. Naveen S)[8], and (GLAUNER)[9] to classify emotions.

SVMs are usually implemented by combining several two-class SVMs. The hyperplane SVMs used in n-dimensional space distinguish points to correct classes by their labels. The optimal hyperplane is the one that is able to form the largest distance between dataset points in this case images. SVMs avoid the "curse of dimensionality" that arises in high-dimensional spaces by tuning the regularization parameter c in linear problems or by kernel trick along with tuning the kernel parameters in nonlinear problems.

The second approach is the adoption of deep learning algorithms, such as CNN, which is the most commonly used deep learning model used for emotion recognition.

Hafiz et al in (Ahamed, Alam and Islam)[10] proposed CNN and HOG as feature extraction in (Ahamed, Alam and Islam)[10], However, in (Li and Deng)[11], Li & Deng used handcrafted HOG features and deep learned features and linear SVM for classification. In (GLAUNER)[9], Patrick also used CNN, and handcrafted features were used for feature extraction. to determine the appropriate feeling for the input image and by reading many articles and studies, finding that the best algorithms used and the most accurate in machine learning to describe the feeling is the algorithm of the SVM and that the best deep learning algorithm in this field is the CNN algorithm, so these two algorithms are used mainly in this study and made some structural improvements and adjustments for both networks to get the best results, and then we did a simple comparison to determine the differences between using deep learning algorithms and machine learning to determine the feeling from the human face.

|  | Network | Features | Dataset | Accuracy |
|---|---|---|---|---|
| (GLAUNER)[9] | CNN<br>Particular smile recognition | Hand-crafted | DISFA | 99.45% |
| (Li and Deng)[11] | ECAN<br><br>Classified seven emotion classes | Deep learning<br>&<br>HOG | JAFFE,<br>MMI,<br>CK+<br>Oulu-CASIA<br>FER2013 SFEW | 61.94%<br>69.89%<br>89.69%<br>63.97%<br>58.21%<br>58.21% |
| (Minaee and Abdolrashidi)[12] | CNN | Hand-crafted | CK+<br>FER2013<br>FERG<br>JAFFE | 98%<br>70.02%<br>99.3%<br>92.8% |
| (Ghaffar)[13] | CNN<br>Classified seven emotion classes |  | JAFFED + KDEF | 78% |
| (Liu, Cheng and Lee)[14] | SVM with genetic algorithm | Geometric feature (landmark curvature, victories landmark) | 8-class<br>CK+<br>7-class<br>CK+<br>7-class<br>MUG | 93.57%<br>95.58%<br>96.29% |
| (Maw, Thu and Mon)[15] | SVM |  | JAFFE | 80% |
| (Alshamsi and Këpuska)[1] | SVM | Facial Landmarks, BRIEF |  | 96.27% |

**Table 1: State of the art**

In the following Table 1, a brief search on the state of the art was done to know the experiments were before and what are the most effective methods to start with.

**Datsets**
The 3RL dataset consists of 24394 images collected and edited from three famous datasets.
As Table 1 shows, the results on CK+ (Extended Cohn-Kanade Dataset) are better than the results on other datasets in terms of accuracy, but it is the smallest dataset between them. "FER-2013", "CK+" and a dataset for facial expression recognition that contains more images to work on, mixing them will lead to a sufficient number of images. However, many misleading images were found in the mentioned datasets, so these images were manually checked one by one, and some deleting, replicating and replacing processes were performed while checking them. A large number of misleading images were noticed that may immediately and negatively affect the output of the applied network. This filtering process has a noticeable improvement on model performance, and balancing between classes was applied to obtain fair results. The resulting dataset (3RL) will be available at https://github.com/Rahma-AZ/3RL-images-dataset-for-Emotion-Recognition-

The three datasets used to create the 3RL dataset are
1- FER-2013 (Facial Expression Recogntion,(FERc), 2013)[16]
2- CK+48 (Cohn-Kanade: (CK+), 2010)[17]
3- A dataset for facial expression recognition (Facial expressions)[18].

Tables 2, 3, and 6 show statistics about previously available datasets.

3.1 FER-2013

|  | Test | Train | sum |
|---|---|---|---|
| Angry | 958 | 3995 | 4953 |
| Disgust | 111 | 436 | 547 |
| Fearful | 1024 | 4097 | 5121 |
| Happy | 1774 | 7215 | 8989 |
| Neutral | 1233 | 4965 | 6198 |
| Sad | 1247 | 4830 | 6077 |
| Surprised | 831 | 3171 | 4002 |
| Sum | 7178 | 28709 | 35888 |

**Table 2: FER-2013 dataset 3.2 CK+48**

|  | Images number |
|---|---|
| Angry | 135 |
| Disgust | 177 |
| Fearful | 75 |
| Happy | 207 |
| Neutral | 54 |
| Sad | 84 |
| Surprised | 249 |
| Sum | 981 |

**Table 3: CK+48 dataset**

3.3 Facial expression recognition dataset
This dataset contains 13718 images collected without classification.

**Dataset Creation Steps**
1) To obtain better results, including the largest number of facial situations, merging datasets may be a good practice. After merging the CK+ 48 dataset, which contains 981 images, with the Fer2013 dataset, which contains 35887 images, the final merged dataset was 36858 images. The dataset was edited manually by removing all misclassified images and some images that contain hands that cover the face completely. The edited merged dataset is shown in table 4.

2) Next, for dataset (CK+ & FER-2013), after deleting all replicated images or images that contain confusion and splitting data, the dataset was as in table 5.

3) A third dataset was added (dataset for facial expression) with the previous datasets (CK+ & FER (2013). This dataset contains 13718 images with (350*350) size, which is available at (Facial expressions)[18], and the dataset was merged and concise from the last seven emotions to five classes (merging between anger & disgust, sad & fear) and almost divided to (80% train and 20% test) with balancing between classes (each class contains approximately the same number of images) in each training class 2000 images and in each test class 400 as in Table 6.

4) Finally, the dataset was maximized by adding some images and replicating others that may contain strong features, which benefit the model to classify better, with a low number of images that include this feature so that each training class had 4000 images, and each test class had 800 images to obtain the 3RL dataset, as shown in table 7.

The whole process explained earlier in all three datasets was concise in the conceptual diagram in Figure 1.
A random collection of photos is shown from the final 3RL dataset in Figure 2.

|  | Test | Train | Sum |
|---|---|---|---|
| Angry | 474 | 3190 | 3664 |
| Disgust | 86 | 399 | 485 |
| Fearful | 493 | 1015 | 1508 |
| Happy | 1547 | 2678 | 4225 |
| Neutral | 805 | 2552 | 3357 |
| Sad | 1134 | 4228 | 5362 |
| Surprised | 504 | 2342 | 2846 |
| Sum | 5043 | 16404 | 21447 |

**Table 4: Merged & Edited dataset**

|  | Test | Train | Sum |
|---|---|---|---|
| Angry | 418 | 1908 | 2326 |
| Disgust | 74 | 299 | 373 |
| Fearful | 233 | 936 | 1169 |
| Happy | 1435 | 5744 | 7179 |
| Neutral | 671 | 2688 | 3359 |
| Sad | 445 | 1782 | 2227 |
| Surprised | 418 | 1676 | 2094 |
| Sum | 3694 | 15033 | 18727 |

Table 5: Cleaned dataset

|  | Test | Train | Sum |
|---|---|---|---|
| Angry | 484 | 2069 | 2553 |
| Happy | 492 | 2039 | 2531 |
| Neutral | 401 | 2000 | 2401 |
| Sad | 403 | 2031 | 2432 |
| Surprised | 400 | 2046 | 2446 |
| Sum | 2180 | 10185 | 12365 |

Table 6: concise dataset

|  | Test | Train | Sum |
|---|---|---|---|
| Angry | 838 | 4040 | 4878 |
| Happy | 822 | 4061 | 4883 |
| Neutral | 854 | 4070 | 4924 |
| Sad | 822 | 4057 | 4879 |
| Surprised | 816 | 4014 | 4830 |
| Sum | 4152 | 20242 | 24394 |

Table 7: 3RL dataset

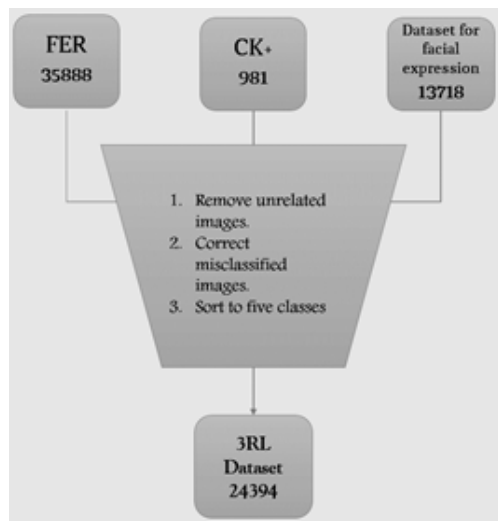

Figure 1: 3RL dataset conceptual diagram

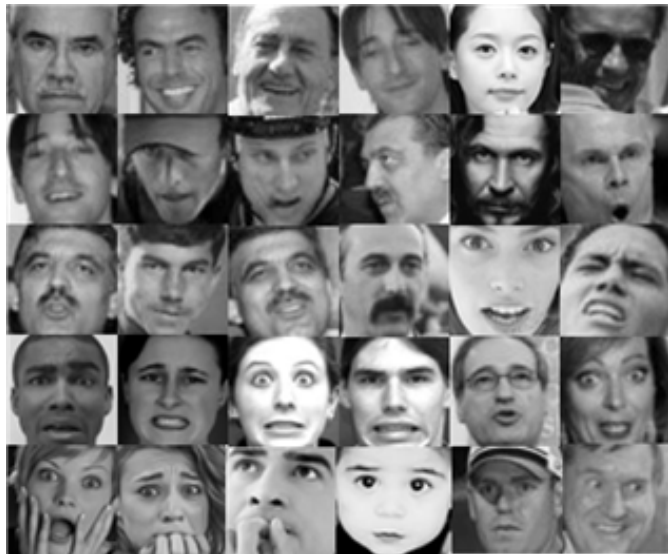

**Figure 2:** 3RL dataset samples

**Setup**

All experiments were performed on core i7 from the seventh generation, 2.8 GHz processor and 16 Giga RAM memory.
OpenCV was used to permit using the camera, and TensorFlow and Keras were used.
Most of the experiments were performed with 50 epochs because a repetitive process was noticed when increasing the number of epochs without obvious differences in accuracy.
In the last experiment in Table 11, 50 epochs of training took approximately seven hours.

**Experiments**
**SVM**

The first experiment on SVM was implemented on FER-2013 as in the Table 2 dataset after extracting landmark and HOG features using 16*16 pixels per cell, and 8 bin pixel gradients were split up into histograms. Using Gridsearch, the best hyperparameters we obtained are as follows: parameter c equals 1, gamma parameter for regularization equals 0.1 with max iterations 10,000 and decision-function one-vs-rest strategy accuracy achieved:

| Kernel | Accuracy | Test Accuracy |
|---|---|---|
| RBF | 99% | 29% |
| Linear | 21% | 14% |
| Sigmoid | 24% | 29% |
| Poly | 32% | 14% |

**Table 8:** Machine FER-2013 result c=1, g=0.1

As shown in table 8, experiments were conducted on different kernels. The best achieved result was obtained by using the radial basis function with an accuracy of 99% and a validation accuracy of 29%.

The same results were found when using the histogram of oriented gradients sliding window instead.
On the 3RL dataset shown in Table 7, better results were achieved with the same decision function, features and kernel radial basis function:

| C | Gamma | Preprocess | Accuracy | Test Accuracy |
|---|---|---|---|---|
| 1 | 1 | ST+PCA136 | 92% | 37% |
| 1 | 0.2 | ST+PCA208 | 99% | 40% |

**Table 9:** Machine Learning Results 3RL

The above Table 9 demonstrates the results when preprocessing the features to avoid the previous overfitting problem.
The extracted features of the 3RL dataset were preprocessed with standard scalars and principal component analysis, which performed feature reduction once with 136 components and obtained 92% accuracy and 37% validation accuracy. Second, 208 components obtained 92% accuracy and 37% validation accuracy. Still not satisfied so, extra attempts were made in Table 10 below without preprocessing and obtained:

| C | Gamma | Accuracy | Test Accuracy |
|---|---|---|---|
| 0.1 | 0.01 | 20% | 20% |
| 1 | 0.1 | 99% | 58% |
| 1 | 0.2 | 99% | 58% |

**Table 10:** Machine Learning Results (without preprocessing)

Table 10 reveals that the final experiments implementing SVM on HOG and landmark features achieved 99% accuracy and 58% validation accuracy.

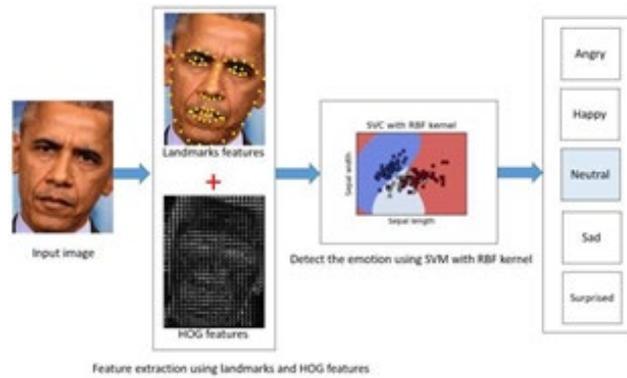

**Figure 3:** ML architecture

Figure 3 shows the ML architecture, where the input image is processed by landmark and HOG feature descriptors. Extracted features are fed to the SVM classifier with radial basis function kernel, decision function one-vs-rest, parameter c with value 1, and parameter gamma assigned 0.2.

| Dataset | Train Accuracy | Test Accuracy |
|---|---|---|
| FER-2013 | 99% | 29% |
| CK+48 | 33% | 32% |
| 3RL dataset | 99% | 58% |

**Table 11: dataset ML comparative**

Table 11 shows the dataset's accuracy with the best achieved parameters for RBF (Radial Basis Function), c, and gamma based on Table 9 and Table 10.

### Deep Learning

First, a similar approach as in (Correa, Jonker, Ozo, & Stolk, 2016)[19], (Chandran and Dr. Naveen S)[8], (Li and Deng)[11], (Zavarez, Berriel, & Oliveira-Santos, 2017)[20] was followed to extract features by CNN, and mainly as (Correa, Jonker, Ozo, & Stolk, 2016)[19] approach was followed and started experiments on FER2013 dataset (Table 2), four layers of convolutional neural network, the filters number in each layer was 32, 64, 128 & 128 for the fourth layer, 64 batch size and 150 epochs with changing the values of network parameters in order to obtain the best results as shown in the Table 12 below:

| LR | Active | Loss FN | Dropout | Accuracy | Test Accuracy |
|---|---|---|---|---|---|
| e-4 | ReLU | categorical | 0.25 0.25 0.5 | 86% | 62% |
| 3e-4 | ReLU | categorical | 0.25 0.5 0.5 | 72% | 62% |
| 3e-4 | Tanh | binary | 0.25 0.5 0.5 | 96% | 92% |

**Table 12: DL (Deep Learning) results 150 epochs, FER-2013**

The last result is considered acceptable, but the prediction results were not satisfactory in all previous experiments. Therefore, a change was made to the merged dataset in Table 4, keeping the learning rate (3e$^{-4}$) and the layers as it was and changing the dropout layers to 0.35, 0.5 and 0.5 for the last layer. Then, the results as shown in Table 13 were obtained:

| Batch | LR | activate | Loss FN | epoch | Accuracy | Test Accuracy |
|---|---|---|---|---|---|---|
| 64 | 3e-4 | ReLU | categorical | 150 | 95% | 66% |
| 32 | 1e-4 | ReLU | Binary | 50 | 88% | 68% |
| 32 | 1e-4 | Tanh | Binary | 100 | 98% | 90% |

**Table 13: Deep Learning Results with Table 4 dataset**

As shown in the previous Table 13, an improvement in the prediction was noticed in the last two experiments when heading to batch size 32, so in the next experiments, the batch size of 32 will be kept.

Another trial was performed using the dataset in Table 5, keeping the other parameters the same as in the last experiment. A 98.9% training accuracy was achieved, and a 92.7% validation accuracy was achieved. The prediction was good but still not enough.

Therefore, the model was trained with the three datasets, adding a dataset for facial expression recognition. The last dataset was cleaned and sorted into five main classes (emotions), as shown in Table 6, and the dropout layers were edited to 0.35, 0.5 & 0.5 for the last one. The results are shown in Table 14.

| Batch | LR | activate | Loss FN | epoch | Accuracy | Test Accuracy |
|---|---|---|---|---|---|---|
| 32 | 1e-4 | ReLU | categorical | 50 | 82% | 66% |

**Table 14: Deep Learning Results table 6 dataset**

For this result in Table 14, the prediction was considered not bad, but reaching a higher accuracy was better, so we concluded that the number of images in the dataset may not be sufficient for deep learn-ing; therefore, the dataset was maximized, as shown in Table 7. In some results with the 3RL dataset, the batch size was 32, the learning rate was 1e-4 and the dropout layers were as in the previous try.

| activate | Loss FN | epoch | Accuracy | Test Accuracy |
|---|---|---|---|---|
| tanh | Binary | 200 | 96% | 80% |
| sigmoid | Categorical | 50 | 90.9% | 90.4% |

**Table 15:** Deep Learning Results with 3RL dataset

In Table 11 the predict was bad so the activation function was changed to ReLU (Rectified Linear Unit), the loss function to binary and the last layer of drop out to 0.25 keeping other parameters as it was, an improvement was noticed in predict, so more experiments were done to obtain a better predict.

Then, tying to change the number of convolutional layer's filters to be 32, 64, 128 and 512 for the last one, the learning rate was e-4 and the batch size was 64, Then, some experiments were done as follow in Table 16:

| Active | Loss FN | Dropout | Accuracy | Test Accuracy |
|---|---|---|---|---|
| ReLU | binary | 0.25 0.1 0.25 | 99.8% | 91% |
| ReLU | categorical | 0.01 0.1 0.25 | 99.5% | 77% |
| ReLU | categorical | 0.01 0.5 0.25 | 99% | 79.5% |
| ReLU | categorical | 0.01 0.03 0.25 | 99.5% | 76% |
| tanh | categorical | 0.01 0.5 0.25 | 97.7% | 79.8% |
| Sigmoid | binary | 0.25 0.1 0.25 | 99.9% | 91.4% |

**Table 16:** Deep Learning Results with different parameters

As noted in Table 16, if the tangent function was used as an activation function with binary cross entropy as a loss function, the result would be more satisfying than using tangent with categorical cross entropy. The best dropout layers were achieved in the first experiment in the previous table, so those dropout layers were retained.

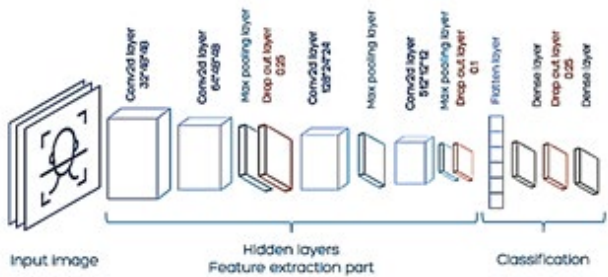

**Figure 4:** Proposed network architecture

The previous figure shows the architecture of the proposed convolutional neural network in the feature extraction and classification parts using the ReLU function as an activation function in each convolution layer, the sigmoid function in the first dense layer and softmax for the last dense layer.

Finally, the desired result was obtained, achieving the best accuracy at 99.9% and 91.4% as validation accuracy, 0.4 for validation loss and 0.004 for loss. The prediction was very good. The network was able to detect all five emotions correctly. Figure 5 shows the curves of accuracy and loss in the last experiment in Table 16.

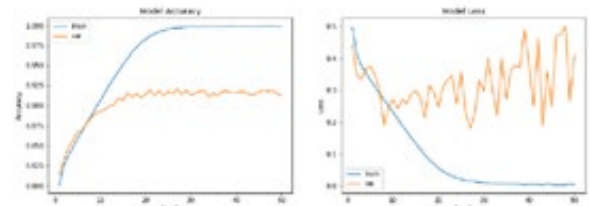

**Figure 5:** Accuracy and loss curves

In Table 17, a short comparison was performed to show the benefit of the 3RL dataset in the prediction phase, where models were tested on FER-2013, CK+48, and 3RL while keeping the same parameters as in the last result in Table 16, changing only the dataset.

| Dataset | Accuracy | Test Accuracy | Prediction |
|---|---|---|---|
| FER-2013 | 99.8% | 89% | Recognize happy, neutral, fear and sad emotion with misclassification in other classes. |
| CK+48 | 99.9% | 95.9% | Only disgust emotion was recognized correctly |
| 3RL dataset | 99.9% | 91.4 | Best prediction results between all experiments |

**Table 17:** datasets DL comparative

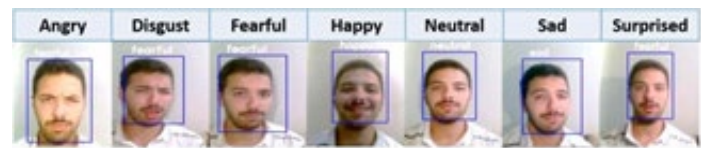

**Figure 6:** FER-2013 predicting results

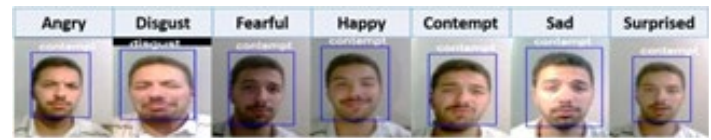

**Figure 7:** CK+48 predicting results

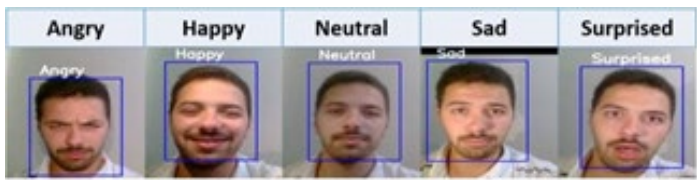

**Figure 8:** 3RL predicting results

Therefore, during experiments, screenshots were taken from real time applying the best parameters as in the last try in Table 16 on the three datasets (FRE-2013, CK+48, 3RL dataset).

A complete disappearance of many emotions was noticed with some errors in prediction, and feeling fear dominated the rest of the feelings when using the FER-2013 dataset, as shown in Figure 6, with the first row of results in Table 17.

When using CK+48 as a dataset, the second row of Table 17 shows the accuracy details. In Figure 7, it is noticeable that most of the emotions are classified as contempt with the absence of a complete classification of other feelings except disgust, which is classified correctly, as the use of this dataset showed a significant decline in the classification compared to the FER-2013 dataset, which gave better results and a noticeable improvement in prediction than CK+48.

In the 3RL dataset, the last row in Table 17, as screenshots of Figure 8, no absence of any feeling from the classification was noted, but very slight and minor errors were noted, such as confusing the feelings of disgust with anger when using the eyebrow-contract feature to express both feelings, which makes this dataset outperform its previous competitors.

## Conclusion & Future Work

Emotion detection is the key to human interaction and understanding. Emotion detection is a challenging task. CNN is one of the best solutions for classifying facial emotions using large datasets such as 3RL. In this study, a new dataset named 3RL was created combining three datasets, CK+48 and Fer2013, and a dataset for facial expression recognition plus merging classes of emotion into 5 main classes. The 3RL dataset will be available at 1. The experiments showed high accuracies of 99%, whereas generalization was better when implementing CNN as the DL method.

The final CNN architecture consists of four 2-D convolutional layers, three dropout layers (32, 64, 128, 512), three max-pooling layers, and two fully connected layers. The input to the network is a preprocessed gray image of 48x48 for the face. The number of layers was selected to maintain a high level of accuracy while still being fast enough for real-time purposes. In addition, max pooling and dropout are utilized more effectively to minimize overfitting.

## Declarations

**Funding**
The authors declare that they have no funding.

## Conflicts of Interest/Competing Interests
The authors declare that they have no competing interests.

## Availability of Data and Material
3RL dataset is available upon request. Code availability Code is available upon request.

## Authors' Contributions
Rahmeh Abou Zafra, Lana Ahmad Abdullah, Rouaa Alaraj, Rasha Albezreh took on the main role so they performed the literature review, conducted the experiments and wrote the manuscript. Tarek Barhoum and Khloud Al Jallad took on a supervisory role. All authors read and approved the final manuscript.

## Consent for Publication
The authors affirm that human research participants provided informed consent for publication of the images in Figure(s) 6, 7 and 8.

## Abbreviations
SVM Support Vector Machine
CNN Convolutional Neural Network
FER Facial Expression Recognition
CK+ Extended Cohn-Kanade Dataset
KNN k Nearest Neighbors
HOG Histogram of Oriented Gradients
RBF Radial Basis Function
DL Deep Learning
ReLU Rectified Linear Unit

## Acknowledgment
This paper and the research behind it would not have been possible without the exceptional support of our supervisors (Dr. Tarek Barhoum and Eng. Khloud Al Jallad). the enthusiasm, knowledge and attention to details throughout our work kept it on track. We also really thank our university (Arab International University) for its support and encouragement to complete this work. We are grateful for our families who supported us and provided total support during doing this research paper. We also thank the proofreader for this research, Miss Salma Al-Hamwi for her efforts to correct the grammar rules in this research paper.